\documentclass[runningheads]{llncs}

\usepackage{eccv}

\usepackage{eccvabbrv}

\usepackage{graphicx}
\usepackage{booktabs}

\usepackage[accsupp]{axessibility}  

\usepackage[breaklinks,colorlinks,citecolor=eccvblue]{hyperref}

\usepackage{orcidlink}

\usepackage[table]{xcolor}
\usepackage{hhline}
\usepackage{float}
\usepackage{booktabs} 
\usepackage{multirow}
\usepackage{fontawesome5}

\begin{document}

\title{OnlineSI: Taming Large Language Model for \\ Online 3D Understanding and Grounding}

\titlerunning{OnlineSI}

\author{Zixian Liu\inst{1} \and
Zhaoxi Chen\inst{2} \and
Liang Pan\textsuperscript{3, \faIcon[regular]{envelope}} \and
Ziwei Liu\textsuperscript{2, \faIcon[regular]{envelope}}}

\authorrunning{Z. Liu et al.}

\institute{Tsinghua University \and
Nanyang Technological University \and
Shanghai AI Lab
}

\maketitle

\begin{figure}[htbp]
    \vspace{-20pt}
    \centering
    \includegraphics[width=\textwidth]{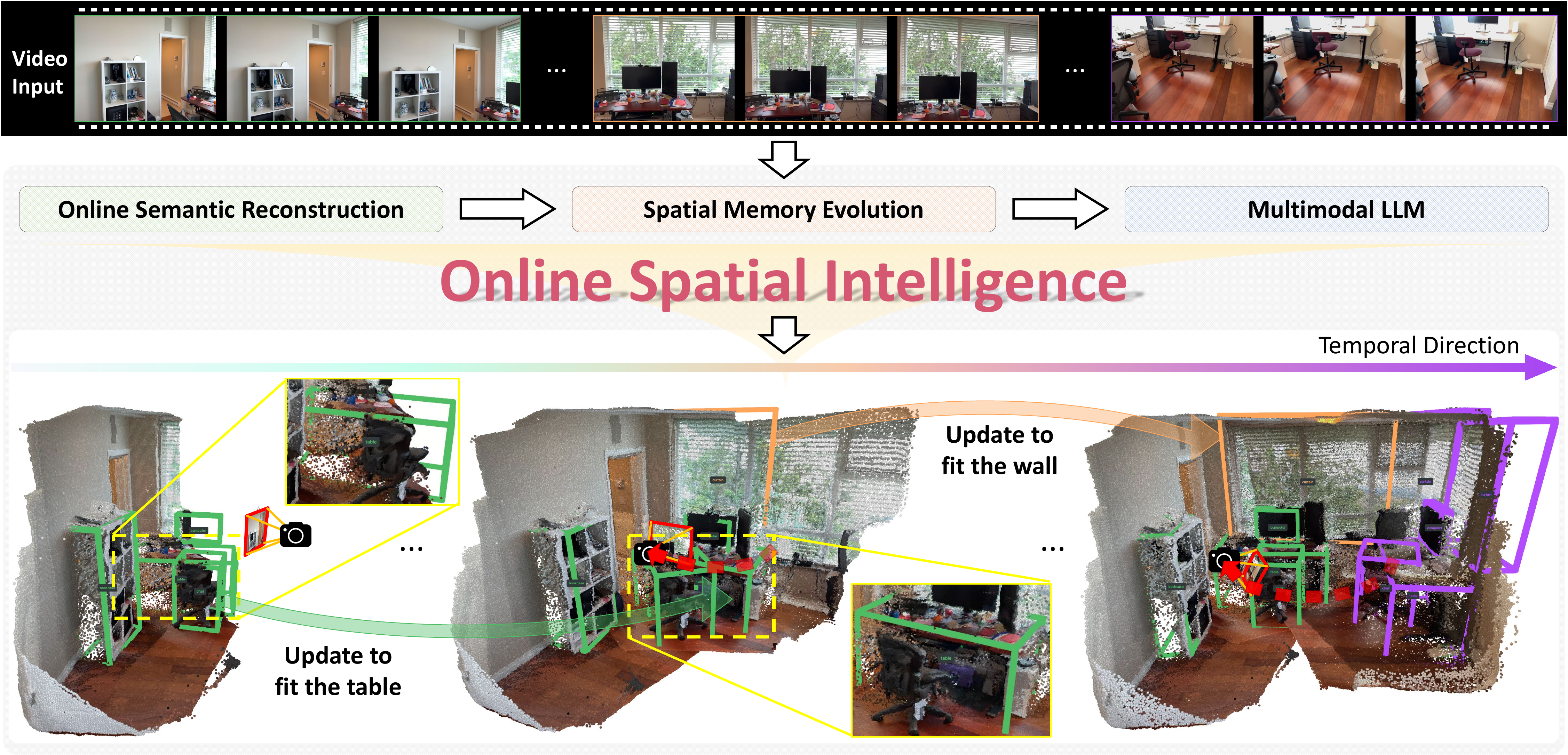}
    \caption{\textbf{OnlineSI} is a framework specifically designed for online 3D understanding and object grounding. Taking a video stream as input, OnlineSI performs incremental semantic reconstruction and leverages a global spatial memory to aggregate observations over time. As demonstrated in the temporal progression, this allows the framework to continuously refine its scene understanding, updating the previous detection results (e.g., ``Update to fit the table'') and incrementally detecting new object instances.}
    \label{fig:teaser}
    \vspace{-20pt}
\end{figure}

\begin{abstract}
  In recent years, researchers have increasingly been interested in how to enable Multimodal Large Language Models (MLLM) to possess spatial understanding and reasoning capabilities.
  However, most existing methods overlook the importance of the ability to continuously work in an ever-changing world, and lack the possibility of deployment on embodied systems in real-world environments.
  In this work, we introduce \textbf{OnlineSI}, a framework that can continuously improve its spatial understanding of its surroundings given a video stream.
  Our core idea is to maintain a finite spatial memory to retain past observations, ensuring the size of the spatial memory does not increase as the input accumulates.
  We further integrate 3D point cloud information with semantic information, helping MLLM to better locate and identify objects in the scene.
  To evaluate our method, we introduce the Fuzzy $F_1$-Score to mitigate ambiguity, and test our method on two representative datasets.
  Experiments demonstrate the effectiveness of our method, paving the way towards real-world embodied systems.
\end{abstract}

\section{Introduction}
\label{sec:introduction}

Processing and understanding streaming information is crucial for embodied systems. 
As robotic agents move from constrained, fixed environments to dynamic, open-world scenes, and as they transition from executing short, simple skills to solving long-horizon tasks, they must be able to continuously learn from an ever-changing 3D world. 
For instance, an autonomous agent exploring a novel environment must constantly update its spatial understanding from new observations and respond on the fly. 
Such online learning capability is a fundamental prerequisite for downstream tasks such as active learning, long-term planning, and human-robot interaction.

However, equipping Multimodal Large Language Models (MLLM) with this form of online spatial intelligence remains challenging. 
First, most existing methods face prohibitive computational scaling as input stream grows. 
The architectures that repeatedly process all past observations with full attention layers quickly exhaust the model context and computational budget due to spatial and contextual redundancy.
Second, many models still fail in understanding spatial relationships and reasoning in space, a limitation that is further exacerbated by the scarcity of 3D-grounded training data. 
While recent works, e.g., Hu et al.~\cite{hu20253dllm}, have explored memory management for embodied planning, their solutions have two key limitations. 
As task execution proceeds, the memory bank infinitely grows, leading to computational and storage bottlenecks. 
Furthermore, their model perceives 3D space at a coarse-grained level, rendering them unsuitable for guiding fine-grained operations such as precise object manipulation. 
These difficulties motivate our core research question: \textit{Can multimodal large language models perceive and understand the 3D world in an online fashion?}

In this paper, we propose \textbf{OnlineSI}, a framework designed to understand 3D scenes in an online fashion and make detections from streaming video.
Our key insight is to maintain a finite, explicit spatial memory, and leverage the multimodal model to reason over this compact memory representation for precise, object-level predictions. 
This design ensures a maximum memory space limit and effectively curbs the growth of the computation budget.
At the same time, the explicit spatial memory structure also facilitates the model's grasp of fine-grained spatial concepts. 
Furthermore, we introduce a novel fusion technique that tightly integrates 3D point cloud data with semantic information, providing the MLLM with a rich, object-level understanding of the scene.

Specifically, for each incoming frame, OnlineSI leverages off-the-shelf models to extract 3D and semantic information, integrating these observations into its global spatial memory. The MLLM then processes this updated memory to output a scene description, including detection results for objects observed. As the scene is progressively reconstructed, the model not only identifies new objects but also automatically refines previous detections that were made under partial observations, as illustrated in Fig.~\ref{fig:teaser}.

Evaluating performance on an arbitrary video stream introduces a significant ambiguity: given the inherently incomplete observation, should a partially viewed object be counted as a detection? To address this evaluation challenge, we propose the Fuzzy $F_1$-Score, a modification of the standard $F_1$-Score designed for this online, partial observation setting. We conduct extensive experiments on two representative datasets, ScanNet~\cite{dai2017scannet} and ScanNet++~\cite{yeshwanthliu2023scannetpp}. Experimental results show that our method significantly outperforms existing baselines, demonstrate its effectiveness in the online setting.

In summary, our contributions are as follows:

\begin{itemize}
    \item We introduce OnlineSI, a novel framework for online 3D scene understanding and grounding that maintains bounded memory space and reduces the growth of inference cost, enabling it to process video streams incrementally.
    \item We propose a new fusion method that integrates 3D point cloud and semantic information, enhancing the MLLM's object-level spatial understanding.
    \item We propose the Fuzzy $F_1$-Score, a new evaluation metric to fairly assess detection performance under the ambiguity of partial observation, and validate our framework's superior performance through extensive experiments.
\end{itemize}
\section{Related Work}
\label{sec:related_work}

\noindent \textbf{3D Object Detection}
aims to predict category labels and oriented 3D bounding boxes for each instance in the scene. Previous work can be divided into two main genres: methods that operate directly on 3D point clouds and those that use single-view 2D images. In the point-cloud-based genre, research has focused on enhancing the model's understanding by enriching the 3D representation. A prominent strategy is to leverage knowledge from other modalities. For instance, \cite{lu2022open, lu2023open, cao2023coda, cao2025collaborative} tap into knowledge priors from 2D images to achieve open-vocabulary 3D detection. Similarly, \cite{peng2024global} employs a Large Language Model (LLM) to better process global-local information. Other approaches improve the model architecture or training strategy. For example, \cite{shen2023v} uses a vertex relative positional encoding method to help the model focus on points near objects, while \cite{kolodiazhnyi2025unidet3d} improves the system's generalization through multi-dataset training. In the single-view-image-based genre, \cite{brazil2023omni3d} pioneers unified monocular 3D object detection by training on the extensive Omni3D dataset. Following this, \cite{li2025towards} introduces the first successful BEV-based monocular 3D object detector. A central theme in this area is addressing the scarcity of 3D training data by leveraging abundant 2D data. To this end, \cite{huang2024training} uses open-vocabulary 2D models and pseudo-LiDAR to automatically label 3D objects. \cite{yao2024open} leverages 2D results from Grounding DINO~\cite{liu2024grounding} with a 3D head to predict 3D bounding boxes. Both \cite{jhang2025v} and \cite{zhang2025detect} also address data scarcity by generating pseudo-3D data from 2D annotations or leveraging pre-trained 2D model knowledge.
In our work, we emphasize on online detection from a video stream. A work with similar topic is~\cite{zhu2025move}, which unifies grounding and exploration for embodied navigation. Compared to their work, we train a large language model to directly understand 3D point clouds and output 3D bounding boxes with rotations and class labels.

\noindent \textbf{Multimodal Large Language Models}
aim to understand, reason, and generate from multimodal information. Early methods, such as \cite{radford2021learning}, unifies images and texts by learning a joint image-text representation. More recently, with the remarkable success of Large Language Model and scaling laws, innumerable works have been devoted to unify vision and language within a single large transformer-based model~\cite{li2022blip, alayrac2022flamingo, li2023blip, liu2023visual, xie2024show, deitke2025molmo, bai2025qwen2, chen2025blip3, diao2025pixels}. Meanwhile, others have attempted to unify multiple modalities, enabling models to understand diverse information sources~\cite{hurst2024gpt, jiang2025solami, xu2025qwen3}.
With the rapid development of autonomous driving and embodied AI, 3D understanding has recently become a significant focus for MLLM researchers. Some works aim to enable models to perceive spatial relationships from images by training on datasets containing 3D annotations~\cite{chen2024spatialvlm, cheng2024spatialrgpt, ma2025spatialllm, xu2025multi}, while others seek to adapt models to 3D representations for spatial reasoning~\cite{fu2024scene, qi2024shapellm, xu2024pointllm, zhu2024llava, fan2025vlm, hu20253dllm, wu2025spatial, zheng2025learning, huang20253d}. However, most of the previous methods suffer from computational overload as the number of input images increases, making them unsuitable for deployment in embodied systems that need to learn at test time. Although some work has explored how to maintain a spatial memory for scene understanding~\cite{yang20253d, zou20253d}, they do not focus on continuously improving understanding and online responding.
In this sense, the work most similar to ours is~\cite{hu20253dllm}. However, they still have problems such as memory bank growing and coarse-level understanding.
In our work, we update an explicit spatial memory based on input frames and use this memory as input to the MLLM, which guarantees a bounded memory space and fine-grained understanding.


\noindent \textbf{2D Object Detection}
is one of the most fundamental and challenging problems in computer vision. In earlier years, researchers typically use carefully designed, handcrafted features for detection~\cite{viola2001rapid, dalal2005histograms, felzenszwalb2008discriminatively}. Then with the rise of deep convolutional neural networks (CNN), many works introduced deep learning to this field, and they can be divided into two groups. The first one follows a ``coarse-to-fine'' process, proposing coarse regions and then focusing on them~\cite{girshick2014rich, he2015spatial, ren2015faster}. The second group, in contrast, retrieves all objects in one-step inference~\cite{redmon2016you, liu2016ssd, lin2017focal, law2018cornernet, carion2020end}. Recently, significant research has focused on open-vocabulary object detection, often achieved through large-scale pre-training~\cite{zang2022open, gu2021open, li2022grounded, liu2024grounding}. In our work, we use an off-the-shelf detector to obtain semantic labels. This semantic information is then injected into the point cloud to enhance its representation.

\section{Preliminaries: SpatialLM}
\label{sec:preliminaries}

We leverage SpatialLM's~\cite{mao2025spatiallm} ability of understanding point clouds to achieve online spatial intelligence. SpatialLM is a multimodal large language model designed to understand 3D indoor environment and generate structured scene descriptions. The model employs Sonata~\cite{wu2025sonata} as its point cloud encoder and finetunes Llama-3.2-1B-Instruct~\cite{touvron2023llama} on a synthetic dataset consisting of 12,328 distinct scenes.

Given the point cloud $\mathbf{P}$ of $N$ points, Sonata first encodes it into 3D feature patches $\mathbf{F}$:

\begin{equation}
    \mathbf{F} = \mathcal{E}(\mathbf{P})\quad \mathbf{P}\in\mathbb{R}^{N\times 6},\ \mathbf{F}\in\mathbb{R}^{K\times D},
\end{equation}

where $\mathbf{K}$ is the number of feature patches and $\mathbf{D}$ is the feature dimension, and each point is represented by coordinates XYZ and color RGB. Then, the feature patches $\mathbf{F}$, along with the prompt tokens, are fed into the large language model backbone, and generate a scene description as text. This output can be subsequently interpreted as a set of 3D bounding boxes for objects in the scene.

It is worth noting that the original SpatialLM model requires the input point cloud to be aligned with the coordinate axes, i.e. the floor must be perpendicular to the z-axis and the walls must be parallel to the x and y-axis. Furthermore, the predicted 3D bounding boxes are constrained to rotations only around the z-axis. This axis-alignment constraint significantly hinders its application in the online detection, where the camera may have an arbitrary 6D pose.
\section{Method}
\label{sec:method}

\begin{figure*}[thbp]
    \vspace{-10pt}
    \centering
    \includegraphics[width=\linewidth]{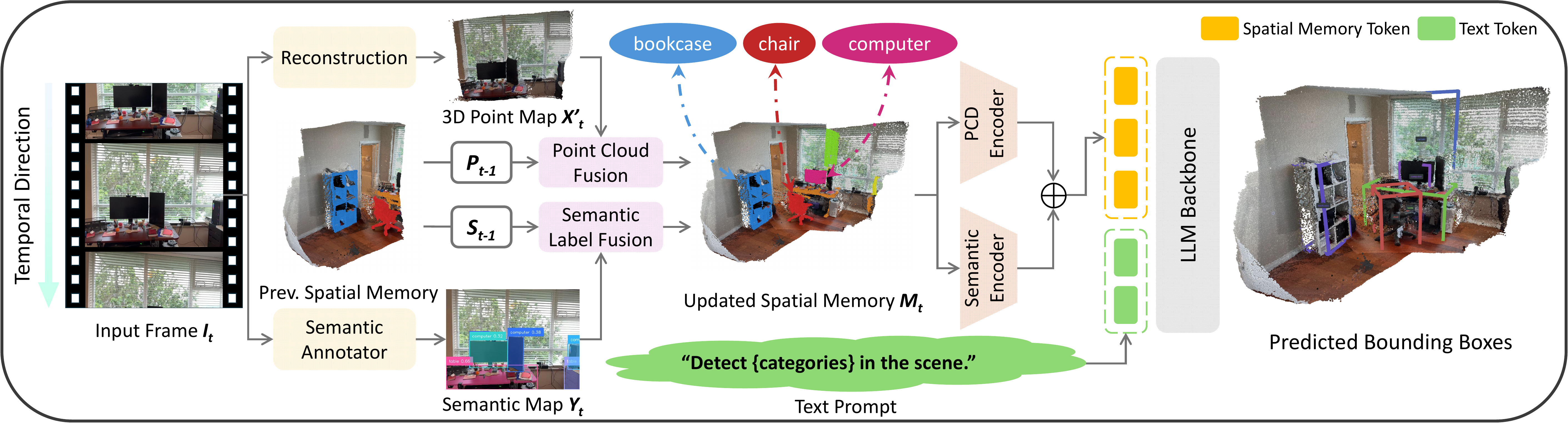}
    \caption{\textbf{Overview of OnlineSI.} For each frame $\mathbf{I}_t$ in the video stream, we first reconstruct the pointmap $\mathbf{X}'_t$, and predict the semantic label for each point $\mathbf{Y}_t$. Next, we fuse the current pointmap and semantic map into the previous spatial memory $\{\mathbf{P}_{t-1}, \mathbf{S}_{t-1}\}$ for updating. We then use the point cloud encoder and the semantic encoder to obtain point cloud features and semantic features, respectively, and add them together as our spatial memory tokens. Lastly, we feed the spatial memory tokens and the text prompt tokens into the LLM backbone, and generate a scene description, which contains the detection results of objects in the current scene.}
    \label{fig:method}
    \vspace{-10pt}
\end{figure*}

We first provide the problem formulation. Then we describe how we maintain the spatial memory from the input frames, and process it using a multimodal large language model. Subsequently, we present the design of the point cloud encoder and the semantic encoder, detailing how semantic information is injected into the point cloud features. Lastly, we introduce the Fuzzy $F_1$-Score, a metric adapted for the online setting. The overview of our framework is demonstrated in Fig.~\ref{fig:method}.

\subsection{Problem Formulation}
\label{sec:problem_formulation}

Given a potentially infinite sequence of images $\{\mathbf{I}_t\in\mathbb{R}^{H\times W\times 3}\}_{t=1}^N$, for each current image $\mathbf{I}_t$, our goal is to predict the 3D bounding boxes and class labels for objects that have appeared in the scene, represented in a unified coordinate frame.
We want the system to be capable of long-term learning from the environment; therefore, the computational and memory consumption of each inference step should not quickly become unbearable as the input accumulates.
A good approach is to maintain a memory $\mathbf{M}_t$ of past observations, and ensure that $\mathbf{M}_t$ has a fixed upper bound on its size.


The multimodal large language model cannot handle point clouds with 3D rotation. Consequently, we define a unified coordinate frame that deviates from the convention used in 3D reconstruction tasks (i.e., the initial camera frame). Instead, our system's origin is at the initial camera's position, but its $xy$ plane is parallel to the ground in the scene and its $z$-axis is perpendicular to it. This coordinate frame is suitable for most embodied agents, where either the camera is looking straight ahead, or we know the camera's relative pose in the world frame. We assume the transformation from the initial camera frame to our unified, aligned coordinate frame is known, which can be derived from the camera's pitch and roll angles. The impact of this coordinate system choice on the final performance is discussed in Sec.~\ref{sec:ablation}.

\subsection{Spatial Memory Management}
\label{sec:memory_management}

We chose to use explicit 3D point clouds to store past observations. The spatial memory $\mathbf{M}_t$ consists of a point cloud part and a semantic part: $\mathbf{M}_t = \{\mathbf{P}_t, \mathbf{S}_t\}$, where $\mathbf{P}_t\in\mathbb{R}^{N\times 6}$ is the constructed point cloud and $\mathbf{S}_t\in\mathbb{Z}^{N}$ is the semantic label of each point. For each frame $\mathbf{I}_t$, we first leverage pre-trained reconstruction model to predict the pointmap $\mathbf{X}_t\in\mathbb{R}^{H\times W\times 3}$ in the initial camera frame, and obtain a semantic map $\mathbf{Y}_t\in\mathbb{Z}^{H\times W}$ through an off-the-shelf semantic annotator, where each pixel in $\mathbf{Y}$ represents the semantic label of corresponding point in $\mathbf{X}_t$. The points in $\mathbf{X}_t$ are then transformed into the unified aligned coordinate frame, yielding $\mathbf{X}’_t$. Subsequently, we fuse $\mathbf{X}'_t$, $\mathbf{Y}_t$ into the previous spatial memory:
\begin{equation}
    \mathbf{P}_t = \text{Fuse}(\mathbf{P}_{t-1}, \mathbf{X}'_t, t),\ \mathbf{S}_t = \text{Fuse}(\mathbf{S}_{t-1}, \mathbf{Y}_t, t),
    \label{eq:memory_management}
\end{equation}
where $\mathbf{P}_{t-1}$, $\mathbf{X}'_t$, $\mathbf{S}_{t-1}$, $\mathbf{Y}_t$ are sampled at specific ratios and then concatenated for updating. The sample indices are unified to ensure that points and their semantic labels correspond. To maintain $\mathbf{M}_t$ at a fixed size and prevent forgetting of early observations as input accumulates, we adjust the fusion ratios based on the timestep $t$. This ensures the total number of points in $\mathbf{M}_t$ remains below a predefined threshold, and the number of points sampled from each input frame is uniform. Regarding the effect of spatial memory management on reducing $\mathbf{M}_t$'s growth, please refer to Sec.~\ref{sec:cost_scaling}.

With the updated spatial memory, we use the point cloud encoder and the semantic encoder to extract point cloud features and semantic features, respectively. We then add them together to create the spatial memory tokens. After that, these spatial memory tokens, along with language instruction tokens, are fed into the large language model to generate a scene description, which contains the 3D bounding boxes of all objects detected thus far in the scene.

\subsection{Point Cloud Encoder and Semantic Encoder}
\label{sec:method_encoders}

Large-scale pre-training enables SpaitalLM to understand various indoor environments. However, we observed that SpatialLM's performance degrades significantly when processing partially reconstructed scenes compared to full scene inputs. Therefore, we inject additional semantic information, which is predicted by pre-trained models, into the point cloud to aid the model in localizing and recognizing objects.

In our framework, the point cloud encoder follows the design of Sonata~\cite{wu2025sonata}. The point cloud $\mathbf{P}$ is encoded into 3D feature patches $\mathbf{F}$, where each patch represents a cubic region. To preserve the locality of point cloud features, our semantic encoder is designed with an identical pooling structure to the point cloud encoder, but it contains no trainable parameters between pooling operations, and only passes through a final linear projection layer for fine-tuning.

Specifically, for the semantic label of each point, we first convert it to its corresponding Llama-3.2-1B-Instruct~\cite{touvron2023llama} token feature. We then use the same pooling operations as the point cloud encoder to aggregate these per-point features into semantic feature patches. After that, we feed the patches into a linear projection layer, and add the resulting features to the point cloud features. This design not only ensures that the semantic information and point cloud features share the same granularity and spatial location, but also minimizes the number of trainable parameters, making fine-tuning possible with limited 3D datasets.

\subsection{Fuzzy $F_1$-Score}
\label{sec:method_fuzzy_score}

Previous 3D bounding box detection works mostly use Mean Average Precision or $F_1$ Score as standard performance metrics. However, with monocular image inputs, occlusions and limited fields of view make it difficult to determine whether an object ``should be detected''. This ambiguity is particularly evident in the online setting, where the camera view can change drastically. For example, if only one leg of a table is visible, the model clearly lacks sufficient information to predict its full dimensions. Conversely, a partially occluded chair in the center of the image should still be detectable.

We propose using Fuzzy $F_1$-Score to mitigate the impact of this ambiguity on the final metrics. Specifically, during evaluation, we define two types of 3D bounding box annotations: the strict ground truth $O^s_{\text{gt}} = \{o^s_{i, \text{gt}}\in\mathbb{R}^7\}_{i=0}^N$ and the lenient ground truth $O^l_{\text{gt}} = \{o^l_{j, \text{gt}}\in\mathbb{R}^7\}_{j=0}^M$, where we use the position of the center, the three dimensions and z-axis rotation to represent a box. The strict ground truth is composed of objects in the lenient ground truth that have a higher visible proportion, hence we have $O^s_{\text{gt}}\subseteq O^l_{\text{gt}}$. These two sets form the lower and upper bounds, respectively, of objects the model is expected to detect. Given the model's predictions $O_{\text{pred}}$, we calculate the $F_1$ score using the recall on $O^s_{\text{gt}}$ and the precision on $O^l_{\text{gt}}$. In other words, let $\text{recall}(O_{\text{pred}}, O^s_{\text{gt}})$ be the recall of $O_{\text{pred}}$ over $O^s_{\text{gt}}$, and $\text{precision}(O_{\text{pred}}, O^l_{\text{gt}})$ be the precision of $O_{\text{pred}}$ over $O^l_{\text{gt}}$, we define:
\begin{equation}
    \text{Fuzzy-$F_1$} = 2\cdot\frac{\text{recall}(O_{\text{pred}}, O^s_{\text{gt}})\cdot\text{precision}(O_{\text{pred}}, O^l_{\text{gt}})}
    {\text{recall}(O_{\text{pred}}, O^s_{\text{gt}}) + \text{precision}(O_{\text{pred}}, O^l_{\text{gt}})},
\end{equation}
The Fuzzy $F_1$-Score is thus unaffected if the model does not detect some objects with low visibility, thereby alleviates the ambiguity problem. For further details on the Fuzzy $F_1$-Score, please refer to Sec.~\ref{sec:ablation}.
\section{Experiments}
\label{sec:experiments}

The primary goal of our experiments is to evaluate our system's ability to continuously build and refine its understanding of novel environments from a monocular video stream. We aim to answer the following research questions: (1) How effectively does our online system perform in novel scenes? (2) How well does our system perform in reducing the growth of inference computation and memory? (3) Is maintaining a persistent spatial memory superior to simply merging independent per-frame predictions? (4) How does the integration of semantic information contribute to overall performance? (5) What is the performance impact of the external modules used in our system?

In this section, we first introduce the implementation details, baselines and datasets. We then present a comprehensive comparison, including both quantitative and qualitative results, to demonstrate the effectiveness of our method. After that, we compare the inference cost scaling of our method with other spatial understanding benchmark, to illustrate how our system reduce the online inference cost. Finally, we present ablation studies, to analyze some key design choices of our method.

\subsection{Experimental Setup}

\noindent \textbf{Implementation Details}
In our system, we choose CUT3R~\cite{wang2025continuous} as our reconstruction module, and Grounded SAM~\cite{ren2024grounded} as our semantic annotator. For the point cloud encoder and the LLM backbone, we initialize them with SpatialLM1.1-Llama-1B~\cite{mao2025spatiallm}. During training, we only set the last layer of the point cloud encoder and the semantic encoder, as well as the LLM backbone as trainable, keeping all other parameters frozen. Training is conducted on 8 NVIDIA A800 GPUs, using standard cross-entropy loss for language model training.

\noindent \textbf{Baselines} We compare our system against a series of baselines:

\begin{itemize}
    \item \textbf{SpatialLM-No-Finetune} We directly feed the reconstructed point cloud into the original SpatialLM model without any training.
    \item \textbf{SpatialLM-Merge} We treat this online process as per-frame prediction. For each frame, we feed the pointmap into SpatialLM to obtain a prediction, then we merge all the detection results to obtain the current results.
    \item \textbf{SpatialLM-Finetune} We maintain the point cloud in the spatial memory but omit the semantic component, leveraging only the point cloud encoder to generate spatial memory tokens.
    \item \textbf{SpatialLM-Ground-Truth-PCD} Identical to SpatialLM-Finetune, but we use ground-truth point clouds instead of reconstructed ones.
    \item \textbf{SpatialLM-Ground-Truth} Identical to our method, but we use ground-truth point clouds and semantic labels.
\end{itemize}

SpatialLM-Ground-Truth-PCD and SpatialLM-Ground-Truth serve as upper-bound benchmarks, illustrating the system's potential performance independent of errors from external modules. Comparisons with the other lower-bound baselines validate the effectiveness of our system's specific design components.

\noindent \textbf{Datasets}
We use ScanNet~\cite{dai2017scannet} and ScanNet++~\cite{yeshwanthliu2023scannetpp} to benchmark our method and baselines. For bounding box annotations, we filter out small objects with all dimensions smaller than 15cm. We define object visibility as the proportion of its visible faces in a given frame and apply different visibility thresholds to filter annotations for training and evaluation. We randomly sample 1$\sim$32 frames with a stride of 30 for each sample sequence.

\subsection{Quantitative Results}

\begin{table}[thbp]
    \vspace{-10pt}
    \centering
    \begin{tabular}{c|c|cccc}
        \toprule
        Method & Avg. Fuzzy-$F_1$ & Chair & Table & Computer & Sink \\
        \midrule
        \rowcolor{eccvblue!20}
        \multicolumn{6}{c}{\textbf{ScanNet++}} \\
        \color{gray}
        SpatialLM-Ground-Truth-PCD & 0.5708 & 0.7538 & 0.6160 & 0.3464 & 0.4078 \\
        \color{gray}
        SpatialLM-Ground-Truth & 0.6420 & 0.7541 & 0.6318 & 0.4540 & 0.6247 \\
        \midrule
        SpatialLM-No-Finetune & 0.0351 & 0.2126 & 0.0000 & 0.0194 & 0.0183 \\
        SpatialLM-Merge & 0.3397 & 0.5357 & 0.4606 & 0.1669 & \textbf{0.2320} \\
        SpatialLM-Finetune & 0.3943 & 0.5732 & \textbf{0.5160} & 0.2036 & 0.1451 \\
        \rowcolor{gray!20}
        \textbf{Ours} & \textbf{0.4397} & \textbf{0.6213} & 0.4717 & \textbf{0.2644} & 0.1683 \\
        \midrule
        \rowcolor{eccvblue!20}
        \multicolumn{6}{c}{\textbf{ScanNet}} \\
        \color{gray}
        SpatialLM-Ground-Truth-PCD & 0.5182 & 0.6050 & 0.5254 & 0.1464 & 0.4511 \\
        \color{gray}
        SpatialLM-Ground-Truth & 0.6138 & 0.6497 & 0.6123 & 0.3059 & 0.6063 \\
        \midrule
        SpatialLM-No-Finetune & 0.0383 & 0.1291 & 0.0000 & 0.0147 & 0.0470 \\
        SpatialLM-Merge & 0.2214 & 0.3267 & 0.2237 & 0.1125 & 0.1964 \\
        SpatialLM-Finetune & 0.2733 & 0.2900 & 0.2893 & 0.1322 & 0.2410 \\
        \rowcolor{gray!20}
        \textbf{Ours} & \textbf{0.3349} & \textbf{0.3425} & \textbf{0.3287} & \textbf{0.1621} & \textbf{0.2712} \\
        \bottomrule
    \end{tabular}
    \caption{\textbf{Quantitative Results of Our Experiments.} We evaluate the methods on ScanNet~\cite{dai2017scannet} and ScanNet++~\cite{yeshwanthliu2023scannetpp}. For each method, we report the average fuzzy $F_1$-score, as well as the fuzzy $F_1$-scores of 4 representative categories. Our method achieves state-of-the-art performance among the non-ground-truth baselines.}
    \label{tab:experiments}
    \vspace{-15pt}
\end{table}

As shown in Tab.~\ref{tab:experiments}, our method significantly outperforms the three lower-bound baselines on two benchmarks. The two upper-bound baselines, which use ground-truth data, establish the potential performance limit of our approach. This result demonstrates the overall effectiveness of our method. Furthermore, our system's superior performance against SpatialLM-Merge highlights the importance of spatial memory maintaining, while the comparison with SpatialLM-Finetune confirms the benefits of injecting semantic information for object localization and recognition.

\subsection{Qualitative Results}

\begin{figure*}[thbp]
    \centering
    \includegraphics[width=\linewidth]{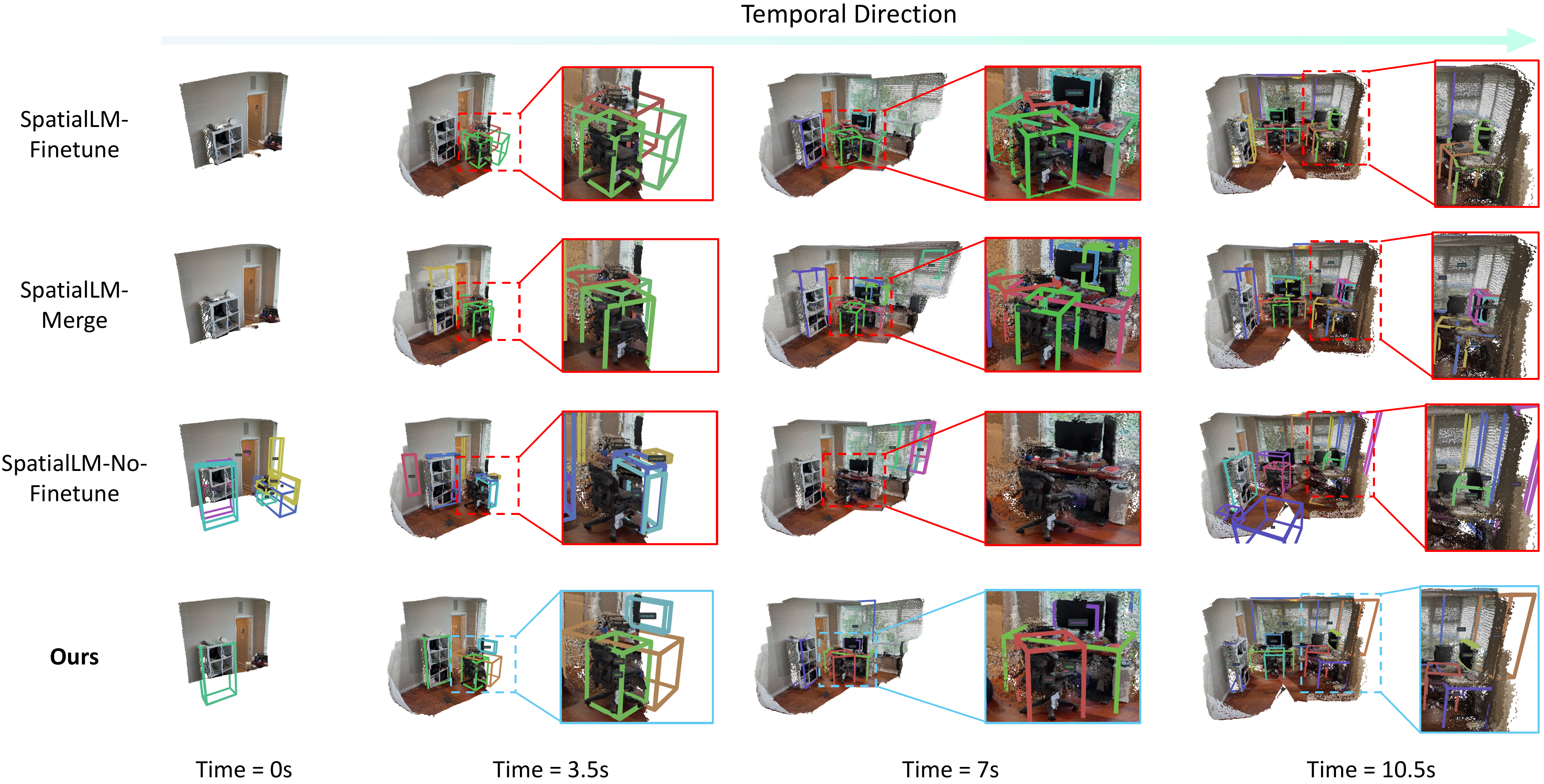}
    \caption{\textbf{Qualitative Results of the Detections.} Here we show examples comparing our method with baselines. We show the progress of scene reconstruction as input frames accumulate, and the detection results at each timestep. We zoom in on key parts of the scene to provide a clearer, more direct comparison.}
    \label{fig:experiments}
    \vspace{-10pt}
\end{figure*}

We compare the prediction results of our method against the three lowe-bound baselines in Fig.~\ref{fig:experiments}. It is evident that the original SpatialLM model (SpatialLM-No-Finetune) performs poorly when fed an online reconstructed point cloud, which is inherently partial and incomplete. Due to the incomplete information in each single-frame prediction, the merged result from SpatialLM-Merge often retains a large number of erroneous predictions. SpatialLM-Finetune often produces incorrect detections due to imperfections in the reconstructed point cloud. In contrast, our method successfully leverages its spatial memory to aggregate information over time and uses semantic cues to assist the model in locating and recognizing objects, enabling the system to build a continuous and accurate understanding of its surroundings in real-world scenes.

\subsection{Inference Cost Scaling}
\label{sec:cost_scaling}

\begin{figure}[htbp]
    \vspace{-10pt}
    \centering
    \includegraphics[width=0.7\linewidth]{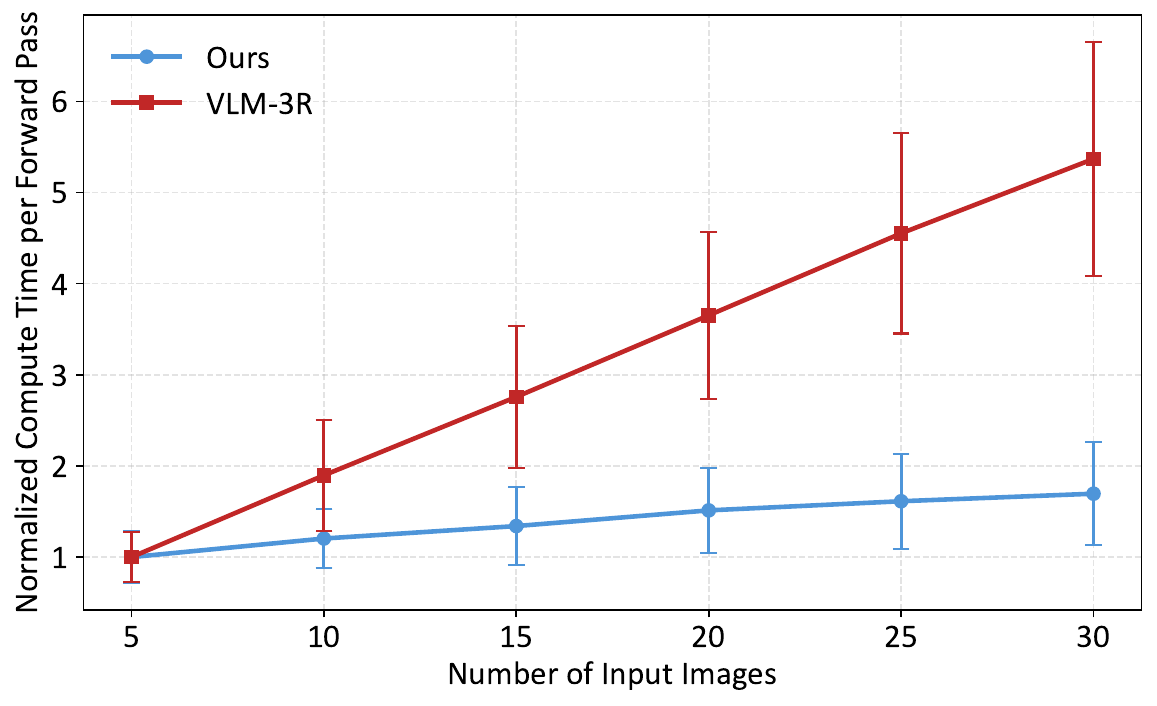}
    \caption{\textbf{Inference Computation Scaling.} Since our method generates different targets than the baseline, we compare the time consumed by a single forward pass of the model rather than the whole generation time. We divide both the computation time and standard deviation by the computation time value at $x=5$ for normalization. All experiments are conducted on the same device.}
    \label{fig:inference_time}
    \vspace{-10pt}
\end{figure}

In this part, we compare the inference cost scaling between our method and VLM-3R~\cite{fan2025vlm} as the number of input images increases. VLM-3R~\cite{fan2025vlm} is similar to our work in utilizing MLLM and CUT3R~\cite{wang2025continuous} for spatial understanding of videos. Due to the different model architectures, we normalized the cost of each method for a fair comparison, ensuring that the two methods were equal at the beginning of scaling. As shown in Fig.~\ref{fig:inference_time}, our method significantly reduces the growth of inference computation compared to other video understanding methods, achieving sub-linear scaling performance.

\begin{figure}[htbp]
    \centering
    \includegraphics[width=0.7\linewidth]{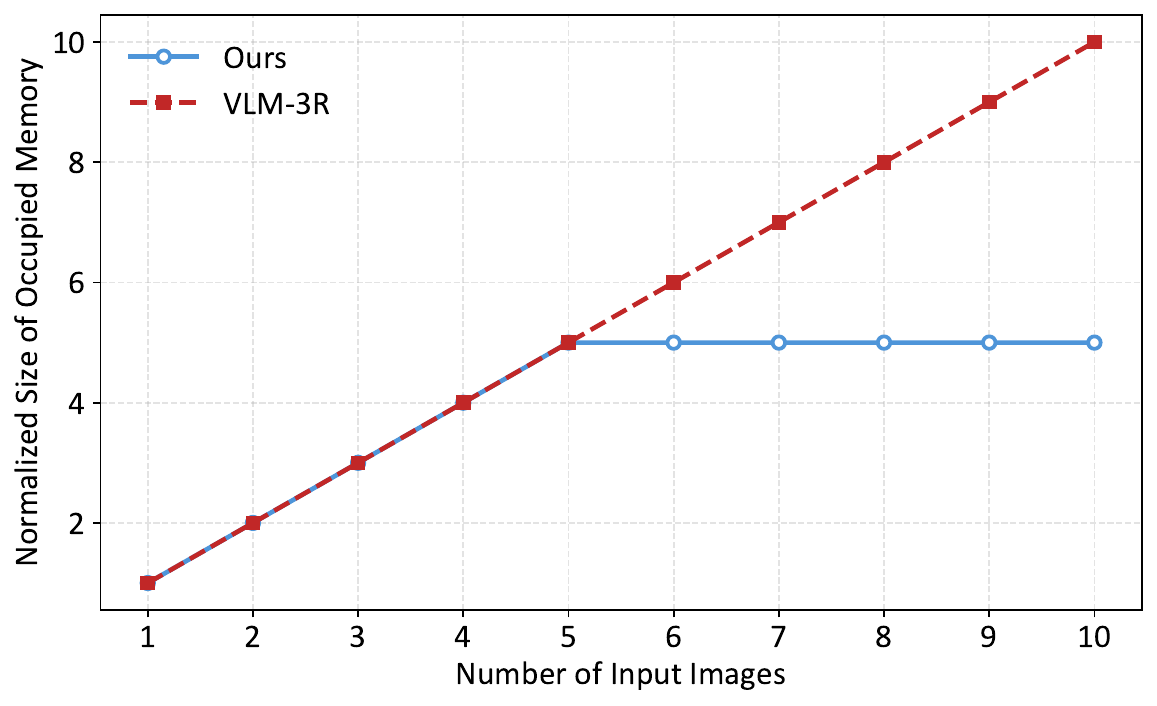}
    \caption{\textbf{Inference Memory Usage Scaling.} Here we use all the previous images as the inference memory of VLM-3R, and use $\mathbf{M}_t$ (refer to Sec.~\ref{sec:memory_management}) as the inference memory of OnlineSI. We divided the memory size by its value at $x=1$ in both methods for normalization.}
    \label{fig:memory_usage}
    \vspace{-10pt}
\end{figure}

Fig.~\ref{fig:memory_usage} illustrates how our system prevents memory usage scaling with spatial memory management. As the baseline continues to grow in terms of memory consumption for inference (because it must store all the past observations), our method remains constant once a preset memory limit is reached. This is because we selectively fuse the past memory with current input, thus avoiding information redundancy.

\subsection{Ablation Study}
\label{sec:ablation}

\begin{table}[htbp]
    \vspace{-10pt}
    \centering
    \begin{tabular}{c|c}
        \toprule
        Memory Representation & Avg. Fuzzy-$F_1$ \\
        \midrule
        1D State & 0.0764 \\
        1D State + Pointmap Feats & 0.0591 \\
        SpatialLM-Fintune & 0.3943 \\
        \bottomrule
    \end{tabular}
    \caption{\textbf{Ablation on the Choice of Spatial Memory.} ``1D State'' means using the latent state representation of CUT3R~\cite{wang2025continuous} as the memory, ``1D State + Pointmap Feats'' means using both the latent state and features in the pointmap output header to obtain memory tokens. Apart from the memory, these two baselines are the same as SpaitalLM-Finetune settings. We conduct experiments on ScanNet++~\cite{yeshwanthliu2023scannetpp} and report the average fuzzy $F_1$-scores.}
    \label{tab:memory_ablation}
    \vspace{-15pt}
\end{table}

\noindent \textbf{Choice of Spatial Memory}
CUT3R~\cite{wang2025continuous} maintains a latent state representation to store the information about the current 3D scene, and the representation in the header before the output pointmap also contains 3D information. Here, we conduct ablation experiments to determine the most suitable spatial memory representation. For these two 1D memory representations we use 5 layers of cross-attention to convert them into memory tokens that are fed into LLM. The results are shown in Tab.~\ref{tab:memory_ablation}. As demonstrated in the results, implicit, one-dimensional memory representations are far less effective than using an explicit three-dimensional point cloud as memory. One possible reason is that the amount of data in ScanNet++ is insufficient for training LLM to understand and interpret these implicit representations.


\begin{figure}[htbp]
    \centering
    \includegraphics[width=0.7\linewidth]{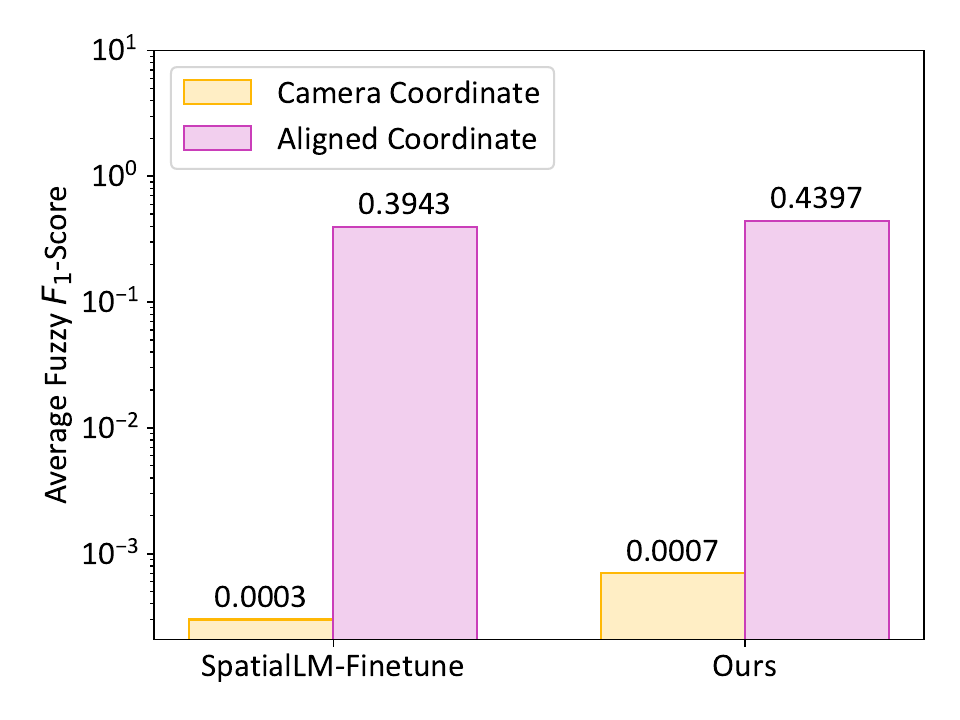}
    \caption{\textbf{Ablation on Coordinate System Selection.} We demonstrate the impact of different coordinate systems of the reconstructed point clouds on performance. We report fuzzy $F_1$-scores evaluated on ScanNet++~\cite{yeshwanthliu2023scannetpp}.}
    \label{fig:coordinate_ablation}
    \vspace{-10pt}
\end{figure}

\noindent \textbf{Coordinate System Selection}
Sec.~\ref{sec:problem_formulation} has demonstrated the importance of choosing correct point cloud coordinate frame. Here we use ablation experiments to quantify the impact of coordinate system selection. As shown in Fig.~\ref{fig:coordinate_ablation}, when we use the initial camera's coordinate frame instead of aligning it with the ground plane, the LLM largely fails to understand the input point cloud. Both our method and SpatialLM-Finetune show a significant drop in performance. This indicates that SpatialLM is incapable of understanding point clouds with arbitrary 3D rotation and can only effectively understand point clouds where the ground plane is aligned, which means the ground plane of the point cloud is parallel to the xy plane of the coordinate frame.

\definecolor{c_pred}{RGB}{14,174,253}
\definecolor{c_anno}{RGB}{146,212,1}
\definecolor{c_s_gt}{RGB}{255,195,3}
\definecolor{c_l_gt}{RGB}{100,100,100}

\begin{figure}[thbp]
    \centering
    \includegraphics[width=\linewidth]{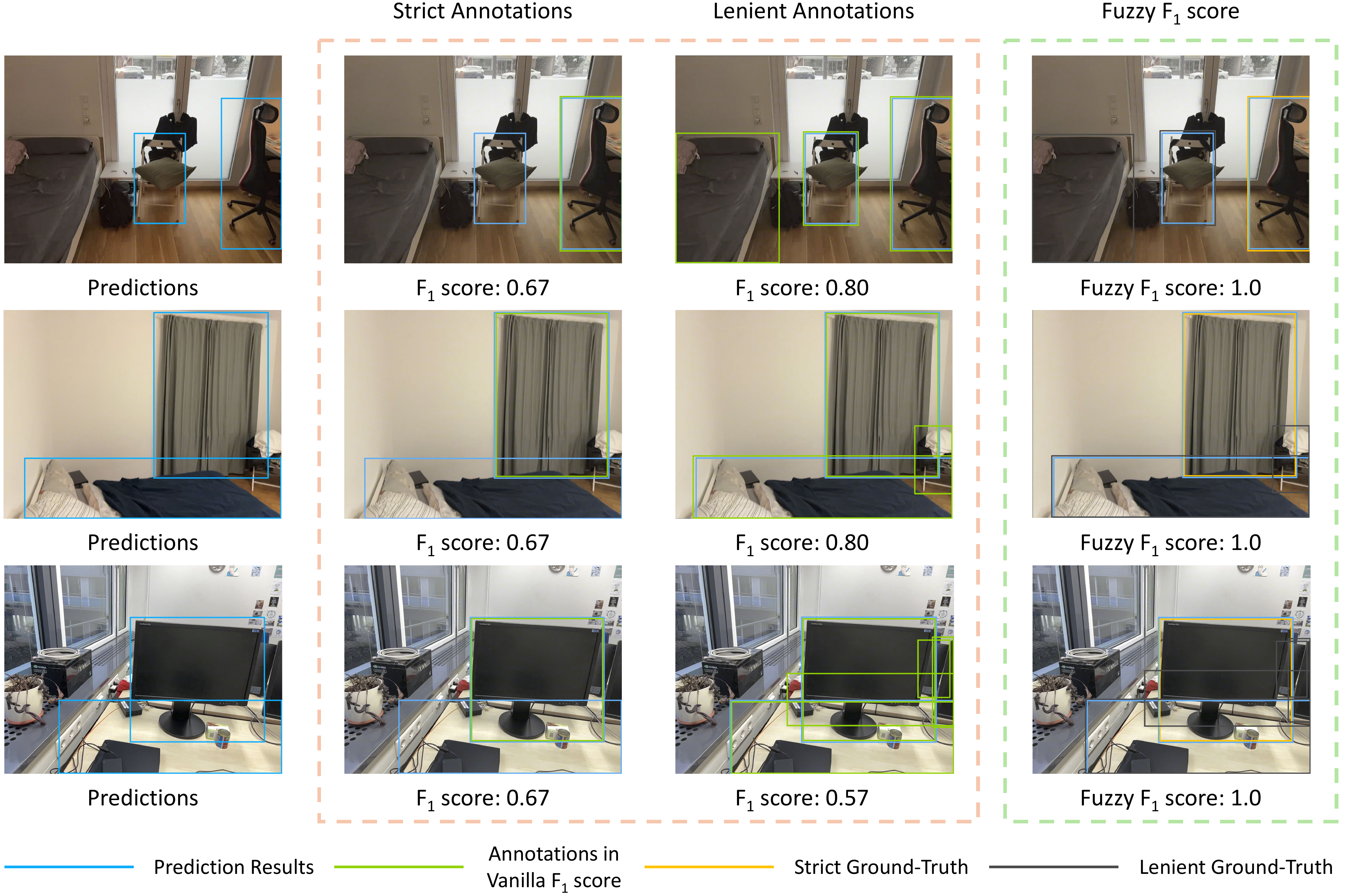}
    \caption{\textbf{Examples of Fuzzy $F_1$-Score.} For each row we show the model's \textcolor{c_pred}{prediction} in the blue boxes. In the second column, we only select objects with high visibility as the evaluation \textcolor{c_anno}{annotation}, while in the third column, we include all low-visibility objects. The vanilla $F_1$-score differs significantly between the two cases. In the last column, we present a visualization of the Fuzzy $F_1$-Score. The \textcolor{c_s_gt}{strict ground-truth} (\textbf{must} be detected) and the \textcolor{c_l_gt}{lenient ground-truth} (\textbf{may} be detected) form the lower and upper bounds of the annotation, providing a more reasonable metrics for evaluation.}
    \label{fig:fuzzy_f1_ablation}
    \vspace{-10pt}
\end{figure}


\noindent \textbf{Fuzzy $F_1$-Score}
Fig.~\ref{fig:fuzzy_f1_ablation} provides some illustrative examples of how our fuzzy $F_1$-score works. \textbf{1.} For the vanilla $F_1$-Score, the final value is greatly affected by different ground truth annotations (see the middle two columns). However, due to partial obstruction, it is difficult to determine, through deterministic procedures, which objects should be detected and used as validating annotation. \textbf{2.} In our Fuzzy $F_1$-Score examples on the last column, we let objects with high visibility form $O^s_{\text{gt}}$, and combine them with low-visibility objects to form $O^l_{\text{gt}}$ (refer to the definition in Sec~\ref{sec:method_fuzzy_score}). Objects in $O^s_{\text{gt}}$ should be detected, while objects in $O^l_{\text{gt}}$ but not in $O^s_{\text{gt}}$ are in an ambiguous state: not detecting them won't affect the final result, correct detection will improve precision, while incorrect detection will decrease precision. This evaluation method ensures that objects which are at the critical point of ``should be detected'' do not affect the final metrics, thus mitigating the ambiguity problem in the online settings.

\begin{table}[htbp]
    \centering
    \begin{tabular}{c|c}
        \toprule
        Semantic Representation & Avg. Fuzzy-$F_1$ \\
        \midrule
        CLIP & 0.4319 \\
        Llama & 0.4397 \\
        \bottomrule
    \end{tabular}
    \caption{\textbf{Ablation on the Semantic Representation.} We experiment with different semantic representations in the semantic encoder. All other settings are the same as our method. We report fuzzy $F_1$-scores evaluated on ScanNet++~\cite{yeshwanthliu2023scannetpp}.}
    \label{tab:semantic_representation_ablation}
    \vspace{-15pt}
\end{table}

\noindent \textbf{Semantic Representation}
In Sec.~\ref{sec:method_encoders} we convert the semantic labels to Llama-3.2-1B-Instruct~\cite{touvron2023llama} token features for semantic information injection. Here we explore the effect of different semantic representation choices. We replace the Llama features to the CLIP~\cite{radford2021learning} features encoded from the same texts in our method, and show the experiments results in Tab.~\ref{tab:semantic_representation_ablation}. As demonstrated in the results, using the same semantic representation as the LLM backbone (Llama) is slightly better than using different semantic representation (CLIP).
\section{Conclusion}
\label{sec:conclusion}

In this work, we introduce a novel framework for online 3D understanding and object grounding. By maintaining a finite spatial memory, our framework can continuously improve its understanding of the environment through the video stream and perform detections in an online fashion. Furthermore, we integrate 3D point cloud data with semantic information, demonstrate how multimodal large language models can understand 3D scenes at a more granular level. To mitigate the ambiguity problem in the online detection, we propose the Fuzzy $F_1$-Score as performance metrics. Experiments demonstrate the effectiveness of our framework.

\noindent \textbf{Limitations}
Despite these contributions, several limitations remain. First, since SpatialLM was pre-trained exclusively on indoor scene datasets, our framework's applicability is currently restricted to similar environments. Enhancing the base model with more diverse 3D data could address this limitation. Second, we currently use a ``sample and then concatenate'' method to maintain the spatial memory, which poses challenges for handling dynamic scenarios. Exploring 4D reconstruction with tracking information to build the spatial memory is an interesting direction. We hope our work could inspire more future work on building spatial intelligence for real-world applications.

%
%
\bibliographystyle{splncs04}
\bibliography{main}

@String(CVPR  = {IEEE Conf. Comput. Vis. Pattern Recog.})

@String(ICCV  = {Int. Conf. Comput. Vis.})

@String(ECCV  = {Eur. Conf. Comput. Vis.})

@String(AAAI  = {AAAI})

@String(CVPR  = {CVPR})

@String(ICCV  = {ICCV})

@String(ECCV  = {ECCV})

@article{lu2022open,
  title={Open-vocabulary 3d detection via image-level class and debiased cross-modal contrastive learning},
  author={Lu, Yuheng and Xu, Chenfeng and Wei, Xiaobao and Xie, Xiaodong and Tomizuka, Masayoshi and Keutzer, Kurt and Zhang, Shanghang},
  journal={arXiv preprint arXiv:2207.01987},
  year={2022}
}

@inproceedings{lu2023open,
  title={Open-vocabulary point-cloud object detection without 3d annotation},
  author={Lu, Yuheng and Xu, Chenfeng and Wei, Xiaobao and Xie, Xiaodong and Tomizuka, Masayoshi and Keutzer, Kurt and Zhang, Shanghang},
  booktitle={Proceedings of the IEEE/CVF conference on computer vision and pattern recognition},
  pages={1190--1199},
  year={2023}
}

@article{shen2023v,
  title={V-detr: Detr with vertex relative position encoding for 3d object detection},
  author={Shen, Yichao and Geng, Zigang and Yuan, Yuhui and Lin, Yutong and Liu, Ze and Wang, Chunyu and Hu, Han and Zheng, Nanning and Guo, Baining},
  journal={arXiv preprint arXiv:2308.04409},
  year={2023}
}

@article{cao2023coda,
  title={Coda: Collaborative novel box discovery and cross-modal alignment for open-vocabulary 3d object detection},
  author={Cao, Yang and Yihan, Zeng and Xu, Hang and Xu, Dan},
  journal={Advances in Neural Information Processing Systems},
  volume={36},
  pages={71862--71873},
  year={2023}
}

@inproceedings{peng2024global,
  title={Global-Local Collaborative Inference with LLM for Lidar-Based Open-Vocabulary Detection},
  author={Peng, Xingyu and Bai, Yan and Gao, Chen and Yang, Lirong and Xia, Fei and Mu, Beipeng and Wang, Xiaofei and Liu, Si},
  booktitle={European Conference on Computer Vision},
  pages={367--384},
  year={2024},
  organization={Springer}
}

@inproceedings{kolodiazhnyi2025unidet3d,
  title={Unidet3d: Multi-dataset indoor 3d object detection},
  author={Kolodiazhnyi, Maksim and Vorontsova, Anna and Skripkin, Matvey and Rukhovich, Danila and Konushin, Anton},
  booktitle={Proceedings of the AAAI Conference on Artificial Intelligence},
  volume={39},
  number={4},
  pages={4365--4373},
  year={2025}
}

@article{cao2025collaborative,
  title={Collaborative novel object discovery and box-guided cross-modal alignment for open-vocabulary 3d object detection},
  author={Cao, Yang and Zeng, Yihan and Xu, Hang and Xu, Dan},
  journal={IEEE Transactions on Pattern Analysis and Machine Intelligence},
  year={2025},
  publisher={IEEE}
}

@inproceedings{brazil2023omni3d,
  title={Omni3d: A large benchmark and model for 3d object detection in the wild},
  author={Brazil, Garrick and Kumar, Abhinav and Straub, Julian and Ravi, Nikhila and Johnson, Justin and Gkioxari, Georgia},
  booktitle={Proceedings of the IEEE/CVF conference on computer vision and pattern recognition},
  pages={13154--13164},
  year={2023}
}

@article{li2025towards,
  title={Towards Unified 3D Object Detection via Algorithm and Data Unification},
  author={Li, Zhuoling and Xu, Xiaogang and Lim, Ser-Nam and Zhao, Hengshuang},
  journal={IEEE Transactions on Pattern Analysis and Machine Intelligence},
  year={2025},
  publisher={IEEE}
}

@article{huang2024training,
  title={Training an open-vocabulary monocular 3d detection model without 3d data},
  author={Huang, Rui and Zheng, Henry and Wang, Yan and Xia, Zhuofan and Pavone, Marco and Huang, Gao},
  journal={Advances in Neural Information Processing Systems},
  volume={37},
  pages={72145--72169},
  year={2024}
}

@article{yao2024open,
  title={Open vocabulary monocular 3d object detection},
  author={Yao, Jin and Gu, Hao and Chen, Xuweiyi and Wang, Jiayun and Cheng, Zezhou},
  journal={arXiv preprint arXiv:2411.16833},
  year={2024}
}

@inproceedings{jhang2025v,
  title={V-mind: Building versatile monocular indoor 3d detector with diverse 2d annotations},
  author={Jhang, Jin-Cheng and Tu, Tao and Wang, Fu-En and Zhang, Ke and Sun, Min and Kuo, Cheng-Hao},
  booktitle={2025 IEEE/CVF Winter Conference on Applications of Computer Vision (WACV)},
  pages={9577--9586},
  year={2025},
  organization={IEEE}
}

@article{zhang2025detect,
  title={Detect anything 3d in the wild},
  author={Zhang, Hanxue and Jiang, Haoran and Yao, Qingsong and Sun, Yanan and Zhang, Renrui and Zhao, Hao and Li, Hongyang and Zhu, Hongzi and Yang, Zetong},
  journal={arXiv preprint arXiv:2504.07958},
  year={2025}
}

@inproceedings{zhu2025move,
  title={Move to understand a 3d scene: Bridging visual grounding and exploration for efficient and versatile embodied navigation},
  author={Zhu, Ziyu and Wang, Xilin and Li, Yixuan and Zhang, Zhuofan and Ma, Xiaojian and Chen, Yixin and Jia, Baoxiong and Liang, Wei and Yu, Qian and Deng, Zhidong and others},
  booktitle={Proceedings of the IEEE/CVF International Conference on Computer Vision},
  pages={8120--8132},
  year={2025}
}

@inproceedings{radford2021learning,
  title={Learning transferable visual models from natural language supervision},
  author={Radford, Alec and Kim, Jong Wook and Hallacy, Chris and Ramesh, Aditya and Goh, Gabriel and Agarwal, Sandhini and Sastry, Girish and Askell, Amanda and Mishkin, Pamela and Clark, Jack and others},
  booktitle={International conference on machine learning},
  pages={8748--8763},
  year={2021},
  organization={PmLR}
}

@inproceedings{li2022blip,
  title={Blip: Bootstrapping language-image pre-training for unified vision-language understanding and generation},
  author={Li, Junnan and Li, Dongxu and Xiong, Caiming and Hoi, Steven},
  booktitle={International conference on machine learning},
  pages={12888--12900},
  year={2022},
  organization={PMLR}
}

@article{alayrac2022flamingo,
  title={Flamingo: a visual language model for few-shot learning},
  author={Alayrac, Jean-Baptiste and Donahue, Jeff and Luc, Pauline and Miech, Antoine and Barr, Iain and Hasson, Yana and Lenc, Karel and Mensch, Arthur and Millican, Katherine and Reynolds, Malcolm and others},
  journal={Advances in neural information processing systems},
  volume={35},
  pages={23716--23736},
  year={2022}
}

@inproceedings{li2023blip,
  title={Blip-2: Bootstrapping language-image pre-training with frozen image encoders and large language models},
  author={Li, Junnan and Li, Dongxu and Savarese, Silvio and Hoi, Steven},
  booktitle={International conference on machine learning},
  pages={19730--19742},
  year={2023},
  organization={PMLR}
}

@article{liu2023visual,
  title={Visual instruction tuning},
  author={Liu, Haotian and Li, Chunyuan and Wu, Qingyang and Lee, Yong Jae},
  journal={Advances in neural information processing systems},
  volume={36},
  pages={34892--34916},
  year={2023}
}

@article{xie2024show,
  title={Show-o: One single transformer to unify multimodal understanding and generation},
  author={Xie, Jinheng and Mao, Weijia and Bai, Zechen and Zhang, David Junhao and Wang, Weihao and Lin, Kevin Qinghong and Gu, Yuchao and Chen, Zhijie and Yang, Zhenheng and Shou, Mike Zheng},
  journal={arXiv preprint arXiv:2408.12528},
  year={2024}
}

@inproceedings{deitke2025molmo,
  title={Molmo and pixmo: Open weights and open data for state-of-the-art vision-language models},
  author={Deitke, Matt and Clark, Christopher and Lee, Sangho and Tripathi, Rohun and Yang, Yue and Park, Jae Sung and Salehi, Mohammadreza and Muennighoff, Niklas and Lo, Kyle and Soldaini, Luca and others},
  booktitle={Proceedings of the Computer Vision and Pattern Recognition Conference},
  pages={91--104},
  year={2025}
}

@article{bai2025qwen2,
  title={Qwen2. 5-vl technical report},
  author={Bai, Shuai and Chen, Keqin and Liu, Xuejing and Wang, Jialin and Ge, Wenbin and Song, Sibo and Dang, Kai and Wang, Peng and Wang, Shijie and Tang, Jun and others},
  journal={arXiv preprint arXiv:2502.13923},
  year={2025}
}

@article{chen2025blip3,
  title={Blip3-o: A family of fully open unified multimodal models-architecture, training and dataset},
  author={Chen, Jiuhai and Xu, Zhiyang and Pan, Xichen and Hu, Yushi and Qin, Can and Goldstein, Tom and Huang, Lifu and Zhou, Tianyi and Xie, Saining and Savarese, Silvio and others},
  journal={arXiv preprint arXiv:2505.09568},
  year={2025}
}

@article{diao2025pixels,
  title={From Pixels to Words--Towards Native Vision-Language Primitives at Scale},
  author={Diao, Haiwen and Li, Mingxuan and Wu, Silei and Dai, Linjun and Wang, Xiaohua and Deng, Hanming and Lu, Lewei and Lin, Dahua and Liu, Ziwei},
  journal={arXiv preprint arXiv:2510.14979},
  year={2025}
}

@article{hurst2024gpt,
  title={Gpt-4o system card},
  author={Hurst, Aaron and Lerer, Adam and Goucher, Adam P and Perelman, Adam and Ramesh, Aditya and Clark, Aidan and Ostrow, AJ and Welihinda, Akila and Hayes, Alan and Radford, Alec and others},
  journal={arXiv preprint arXiv:2410.21276},
  year={2024}
}

@inproceedings{jiang2025solami,
  title={Solami: Social vision-language-action modeling for immersive interaction with 3d autonomous characters},
  author={Jiang, Jianping and Xiao, Weiye and Lin, Zhengyu and Zhang, Huaizhong and Ren, Tianxiang and Gao, Yang and Lin, Zhiqian and Cai, Zhongang and Yang, Lei and Liu, Ziwei},
  booktitle={Proceedings of the Computer Vision and Pattern Recognition Conference},
  pages={26887--26898},
  year={2025}
}

@article{xu2025qwen3,
  title={Qwen3-omni technical report},
  author={Xu, Jin and Guo, Zhifang and Hu, Hangrui and Chu, Yunfei and Wang, Xiong and He, Jinzheng and Wang, Yuxuan and Shi, Xian and He, Ting and Zhu, Xinfa and others},
  journal={arXiv preprint arXiv:2509.17765},
  year={2025}
}

@inproceedings{chen2024spatialvlm,
  title={Spatialvlm: Endowing vision-language models with spatial reasoning capabilities},
  author={Chen, Boyuan and Xu, Zhuo and Kirmani, Sean and Ichter, Brain and Sadigh, Dorsa and Guibas, Leonidas and Xia, Fei},
  booktitle={Proceedings of the IEEE/CVF Conference on Computer Vision and Pattern Recognition},
  pages={14455--14465},
  year={2024}
}

@article{cheng2024spatialrgpt,
  title={Spatialrgpt: Grounded spatial reasoning in vision-language models},
  author={Cheng, An-Chieh and Yin, Hongxu and Fu, Yang and Guo, Qiushan and Yang, Ruihan and Kautz, Jan and Wang, Xiaolong and Liu, Sifei},
  journal={Advances in Neural Information Processing Systems},
  volume={37},
  pages={135062--135093},
  year={2024}
}

@inproceedings{ma2025spatialllm,
  title={Spatialllm: A compound 3d-informed design towards spatially-intelligent large multimodal models},
  author={Ma, Wufei and Ye, Luoxin and de Melo, Celso M and Yuille, Alan and Chen, Jieneng},
  booktitle={Proceedings of the Computer Vision and Pattern Recognition Conference},
  pages={17249--17260},
  year={2025}
}

@article{xu2025multi,
  title={Multi-spatialmllm: Multi-frame spatial understanding with multi-modal large language models},
  author={Xu, Runsen and Wang, Weiyao and Tang, Hao and Chen, Xingyu and Wang, Xiaodong and Chu, Fu-Jen and Lin, Dahua and Feiszli, Matt and Liang, Kevin J},
  journal={arXiv preprint arXiv:2505.17015},
  year={2025}
}

@article{fu2024scene,
  title={Scene-llm: Extending language model for 3d visual understanding and reasoning},
  author={Fu, Rao and Liu, Jingyu and Chen, Xilun and Nie, Yixin and Xiong, Wenhan},
  journal={arXiv preprint arXiv:2403.11401},
  year={2024}
}

@inproceedings{qi2024shapellm,
  title={Shapellm: Universal 3d object understanding for embodied interaction},
  author={Qi, Zekun and Dong, Runpei and Zhang, Shaochen and Geng, Haoran and Han, Chunrui and Ge, Zheng and Yi, Li and Ma, Kaisheng},
  booktitle={European Conference on Computer Vision},
  pages={214--238},
  year={2024},
  organization={Springer}
}

@inproceedings{xu2024pointllm,
  title={Pointllm: Empowering large language models to understand point clouds},
  author={Xu, Runsen and Wang, Xiaolong and Wang, Tai and Chen, Yilun and Pang, Jiangmiao and Lin, Dahua},
  booktitle={European Conference on Computer Vision},
  pages={131--147},
  year={2024},
  organization={Springer}
}

@article{zhu2024llava,
  title={Llava-3d: A simple yet effective pathway to empowering lmms with 3d-awareness},
  author={Zhu, Chenming and Wang, Tai and Zhang, Wenwei and Pang, Jiangmiao and Liu, Xihui},
  journal={arXiv preprint arXiv:2409.18125},
  year={2024}
}

@article{fan2025vlm,
  title={VLM-3R: Vision-Language Models Augmented with Instruction-Aligned 3D Reconstruction},
  author={Fan, Zhiwen and Zhang, Jian and Li, Renjie and Zhang, Junge and Chen, Runjin and Hu, Hezhen and Wang, Kevin and Qu, Huaizhi and Wang, Dilin and Yan, Zhicheng and others},
  journal={arXiv preprint arXiv:2505.20279},
  year={2025}
}

@article{hu20253dllm,
  title={3DLLM-Mem: Long-Term Spatial-Temporal Memory for Embodied 3D Large Language Model},
  author={Hu, Wenbo and Hong, Yining and Wang, Yanjun and Gao, Leison and Wei, Zibu and Yao, Xingcheng and Peng, Nanyun and Bitton, Yonatan and Szpektor, Idan and Chang, Kai-Wei},
  journal={arXiv preprint arXiv:2505.22657},
  year={2025}
}

@article{wu2025spatial,
  title={Spatial-mllm: Boosting mllm capabilities in visual-based spatial intelligence},
  author={Wu, Diankun and Liu, Fangfu and Hung, Yi-Hsin and Duan, Yueqi},
  journal={arXiv preprint arXiv:2505.23747},
  year={2025}
}

@article{zheng2025learning,
  title={Learning from Videos for 3D World: Enhancing MLLMs with 3D Vision Geometry Priors},
  author={Zheng, Duo and Huang, Shijia and Li, Yanyang and Wang, Liwei},
  journal={arXiv preprint arXiv:2505.24625},
  year={2025}
}

@article{huang20253d,
  title={3d-r1: Enhancing reasoning in 3d vlms for unified scene understanding},
  author={Huang, Ting and Zhang, Zeyu and Tang, Hao},
  journal={arXiv preprint arXiv:2507.23478},
  year={2025}
}

@inproceedings{yang20253d,
  title={3D-mem: 3D scene memory for embodied exploration and reasoning},
  author={Yang, Yuncong and Yang, Han and Zhou, Jiachen and Chen, Peihao and Zhang, Hongxin and Du, Yilun and Gan, Chuang},
  booktitle={Proceedings of the Computer Vision and Pattern Recognition Conference},
  pages={17294--17303},
  year={2025}
}

@inproceedings{zou20253d,
  title={3D-SPATIAL MULTIMODAL MEMORY},
  author={Zou, Xueyan and Song, Yuchen and Qiu, Ri-Zhao and Peng, Xuanbin and Ye, Jianglong and Liu, Sifei and Wang, Xiaolong},
  booktitle={The Thirteenth International Conference on Learning Representations},
  year={2025}
}

@inproceedings{viola2001rapid,
  title={Rapid object detection using a boosted cascade of simple features},
  author={Viola, Paul and Jones, Michael},
  booktitle={Proceedings of the 2001 IEEE computer society conference on computer vision and pattern recognition. CVPR 2001},
  volume={1},
  pages={I--I},
  year={2001},
  organization={Ieee}
}

@inproceedings{dalal2005histograms,
  title={Histograms of oriented gradients for human detection},
  author={Dalal, Navneet and Triggs, Bill},
  booktitle={2005 IEEE computer society conference on computer vision and pattern recognition (CVPR'05)},
  volume={1},
  pages={886--893},
  year={2005},
  organization={Ieee}
}

@inproceedings{felzenszwalb2008discriminatively,
  title={A discriminatively trained, multiscale, deformable part model},
  author={Felzenszwalb, Pedro and McAllester, David and Ramanan, Deva},
  booktitle={2008 IEEE conference on computer vision and pattern recognition},
  pages={1--8},
  year={2008},
  organization={Ieee}
}

@inproceedings{girshick2014rich,
  title={Rich feature hierarchies for accurate object detection and semantic segmentation},
  author={Girshick, Ross and Donahue, Jeff and Darrell, Trevor and Malik, Jitendra},
  booktitle={Proceedings of the IEEE conference on computer vision and pattern recognition},
  pages={580--587},
  year={2014}
}

@article{he2015spatial,
  title={Spatial pyramid pooling in deep convolutional networks for visual recognition},
  author={He, Kaiming and Zhang, Xiangyu and Ren, Shaoqing and Sun, Jian},
  journal={IEEE transactions on pattern analysis and machine intelligence},
  volume={37},
  number={9},
  pages={1904--1916},
  year={2015},
  publisher={IEEE}
}

@article{ren2015faster,
  title={Faster r-cnn: Towards real-time object detection with region proposal networks},
  author={Ren, Shaoqing and He, Kaiming and Girshick, Ross and Sun, Jian},
  journal={Advances in neural information processing systems},
  volume={28},
  year={2015}
}

@inproceedings{redmon2016you,
  title={You only look once: Unified, real-time object detection},
  author={Redmon, Joseph and Divvala, Santosh and Girshick, Ross and Farhadi, Ali},
  booktitle={Proceedings of the IEEE conference on computer vision and pattern recognition},
  pages={779--788},
  year={2016}
}

@inproceedings{liu2016ssd,
  title={Ssd: Single shot multibox detector},
  author={Liu, Wei and Anguelov, Dragomir and Erhan, Dumitru and Szegedy, Christian and Reed, Scott and Fu, Cheng-Yang and Berg, Alexander C},
  booktitle={European conference on computer vision},
  pages={21--37},
  year={2016},
  organization={Springer}
}

@inproceedings{lin2017focal,
  title={Focal loss for dense object detection},
  author={Lin, Tsung-Yi and Goyal, Priya and Girshick, Ross and He, Kaiming and Doll{\'a}r, Piotr},
  booktitle={Proceedings of the IEEE international conference on computer vision},
  pages={2980--2988},
  year={2017}
}

@inproceedings{law2018cornernet,
  title={Cornernet: Detecting objects as paired keypoints},
  author={Law, Hei and Deng, Jia},
  booktitle={Proceedings of the European conference on computer vision (ECCV)},
  pages={734--750},
  year={2018}
}

@inproceedings{carion2020end,
  title={End-to-end object detection with transformers},
  author={Carion, Nicolas and Massa, Francisco and Synnaeve, Gabriel and Usunier, Nicolas and Kirillov, Alexander and Zagoruyko, Sergey},
  booktitle={European conference on computer vision},
  pages={213--229},
  year={2020},
  organization={Springer}
}

@inproceedings{zang2022open,
  title={Open-vocabulary detr with conditional matching},
  author={Zang, Yuhang and Li, Wei and Zhou, Kaiyang and Huang, Chen and Loy, Chen Change},
  booktitle={European conference on computer vision},
  pages={106--122},
  year={2022},
  organization={Springer}
}

@article{gu2021open,
  title={Open-vocabulary object detection via vision and language knowledge distillation},
  author={Gu, Xiuye and Lin, Tsung-Yi and Kuo, Weicheng and Cui, Yin},
  journal={arXiv preprint arXiv:2104.13921},
  year={2021}
}

@inproceedings{li2022grounded,
  title={Grounded language-image pre-training},
  author={Li, Liunian Harold and Zhang, Pengchuan and Zhang, Haotian and Yang, Jianwei and Li, Chunyuan and Zhong, Yiwu and Wang, Lijuan and Yuan, Lu and Zhang, Lei and Hwang, Jenq-Neng and others},
  booktitle={Proceedings of the IEEE/CVF conference on computer vision and pattern recognition},
  pages={10965--10975},
  year={2022}
}

@article{mao2025spatiallm,
  title={SpatialLM: Training Large Language Models for Structured Indoor Modeling},
  author={Mao, Yongsen and Zhong, Junhao and Fang, Chuan and Zheng, Jia and Tang, Rui and Zhu, Hao and Tan, Ping and Zhou, Zihan},
  journal={arXiv preprint arXiv:2506.07491},
  year={2025}
}

@inproceedings{wang2025continuous,
  title={Continuous 3d perception model with persistent state},
  author={Wang, Qianqian and Zhang, Yifei and Holynski, Aleksander and Efros, Alexei A and Kanazawa, Angjoo},
  booktitle={Proceedings of the Computer Vision and Pattern Recognition Conference},
  pages={10510--10522},
  year={2025}
}

@inproceedings{wu2025sonata,
  title={Sonata: Self-supervised learning of reliable point representations},
  author={Wu, Xiaoyang and DeTone, Daniel and Frost, Duncan and Shen, Tianwei and Xie, Chris and Yang, Nan and Engel, Jakob and Newcombe, Richard and Zhao, Hengshuang and Straub, Julian},
  booktitle={Proceedings of the Computer Vision and Pattern Recognition Conference},
  pages={22193--22204},
  year={2025}
}

@article{touvron2023llama,
  title={Llama: Open and efficient foundation language models},
  author={Touvron, Hugo and Lavril, Thibaut and Izacard, Gautier and Martinet, Xavier and Lachaux, Marie-Anne and Lacroix, Timoth{\'e}e and Rozi{\`e}re, Baptiste and Goyal, Naman and Hambro, Eric and Azhar, Faisal and others},
  journal={arXiv preprint arXiv:2302.13971},
  year={2023}
}

@inproceedings{liu2024grounding,
  title={Grounding dino: Marrying dino with grounded pre-training for open-set object detection},
  author={Liu, Shilong and Zeng, Zhaoyang and Ren, Tianhe and Li, Feng and Zhang, Hao and Yang, Jie and Jiang, Qing and Li, Chunyuan and Yang, Jianwei and Su, Hang and others},
  booktitle={European conference on computer vision},
  pages={38--55},
  year={2024},
  organization={Springer}
}

@misc{ren2024grounded,
      title={Grounded SAM: Assembling Open-World Models for Diverse Visual Tasks}, 
      author={Tianhe Ren and Shilong Liu and Ailing Zeng and Jing Lin and Kunchang Li and He Cao and Jiayu Chen and Xinyu Huang and Yukang Chen and Feng Yan and Zhaoyang Zeng and Hao Zhang and Feng Li and Jie Yang and Hongyang Li and Qing Jiang and Lei Zhang},
      year={2024},
      eprint={2401.14159},
      archivePrefix={arXiv},
      primaryClass={cs.CV}
}

@inproceedings{dai2017scannet,
    title={ScanNet: Richly-annotated 3D Reconstructions of Indoor Scenes},
    author={Dai, Angela and Chang, Angel X. and Savva, Manolis and Halber, Maciej and Funkhouser, Thomas and Nie{\ss}ner, Matthias},
    booktitle = {Proc. Computer Vision and Pattern Recognition (CVPR), IEEE},
    year = {2017}
}

@inproceedings{yeshwanthliu2023scannetpp,
  title={ScanNet++: A High-Fidelity Dataset of 3D Indoor Scenes},
  author={Yeshwanth, Chandan and Liu, Yueh-Cheng and Nie{\ss}ner, Matthias and Dai, Angela},
  booktitle = {Proceedings of the International Conference on Computer Vision ({ICCV})},
  year={2023}
}

\end{document}